\documentclass[10pt,twocolumn,letterpaper]{article}

\usepackage{cvpr}              % To produce the CAMERA-READY version
\usepackage[dvipsnames]{xcolor}

\usepackage{algorithm}
\usepackage{listings}

\definecolor{cvprblue}{rgb}{0.21,0.49,0.74}
\usepackage[pagebackref,breaklinks,colorlinks,citecolor=cvprblue]{hyperref}

\usepackage[accsupp]{axessibility}
\usepackage{multirow}
\usepackage{subcaption}
\usepackage{tabularx}

\def\paperID{5591} % *** Enter the Paper ID here
\def\confName{CVPR}
\def\confYear{2024}

\title{Transfer CLIP for Generalizable Image Denoising}

\author{Jun Cheng, Dong Liang, Shan Tan\footnotemark[1]\\
Huazhong University of Science and Technology, Wuhan, China\\
{\tt\small \{jcheng24, liangdong, shantan\}@hust.edu.cn}
}

\begin{document}
\maketitle

\renewcommand{\thefootnote}{\fnsymbol{footnote}}
\footnotetext[1]{Corresponding author.}

\begin{abstract}

Image denoising is a fundamental task in computer vision. While prevailing deep learning-based supervised and self-supervised methods have excelled in eliminating in-distribution noise, their susceptibility to out-of-distribution (OOD) noise remains a significant challenge. The recent emergence of contrastive language-image pre-training (CLIP) model has showcased exceptional capabilities in open-world image recognition and segmentation. Yet, the potential for leveraging CLIP to enhance the robustness of low-level tasks remains largely unexplored. This paper uncovers that certain dense features extracted from the frozen ResNet image encoder of CLIP exhibit distortion-invariant and content-related properties, which are highly desirable for generalizable denoising. Leveraging these properties, we devise an asymmetrical encoder-decoder denoising network, which incorporates dense features including the noisy image and its multi-scale features from the frozen ResNet encoder of CLIP into a learnable image decoder to achieve generalizable denoising. The progressive feature augmentation strategy is further proposed to mitigate feature overfitting and improve the robustness of the learnable decoder. Extensive experiments and comparisons conducted across diverse OOD noises, including synthetic noise, real-world sRGB noise, and low-dose CT image noise, demonstrate the superior generalization ability of our method.

\end{abstract}    
\section{Introduction}

Image denoising is a significant task in computer vision and image processing. Current supervised denoising methods leveraging powerful deep neural networks and large-scale datasets have achieved exceptional performance in both synthetic and real-world noise removal \cite{Restormer,StripTransformer}. However, these supervised denoisers tend to overfit the noise present in the training datasets, resulting in poor generalization to out-of-distribution (OOD) noise \cite{maskdenoising}. On the other hand, unsupervised and self-supervised denoising methods \cite{Noise2void,LIRNet, Neighbor2neighbor,APBSN,CVF-SID,LGBPN} directly focus on the target domain in which the target noisy images reside and hence bypass OOD generalization. Nevertheless, these methods are inherently vulnerable to unseen noise \cite{MeD} and the collection of target noisy datasets is not always available. Therefore, it is critical to enhance the generalization of deep denoisers. 

OOD generalization has been popular research in high-level vision tasks like image recognition and segmentation \cite{OOD_survey1, OOD_survey2}. In contrast, attention to OOD generalization within image denoising is limited. Existing research in this area primarily consists of two aspects: generalization across degradation levels and generalization across degradation types. Regarding the former, some works trained blind denoisers \cite{DNCNN,Restormer,Neighbor2neighbor} or bias-free networks \cite{BFCNN,DRUNet} to handle noise with varying levels. However, these methods are confined to specific noise and cannot generalize to unseen noise types. For the latter, several works aimed to fortify models against general OOD noise. Particularly, MaskDenoising \cite{maskdenoising} incorporated dropout units into the model training to enforce the denoiser to learn the reconstruction of image contents. DIL \cite{DIL} built upon causality and meta-learning and encouraged the model to learn distortion-invariant representations. HAT \cite{HAT} designed an adversarial attack for deep denoisers and then conducted adversarial training. 

Recently, through solving the image-text alignment problem based on hyper-scale datasets, the contrastive language-image pre-training (CLIP) model \cite{CLIP} has demonstrated remarkable generalization capacity in downstream open-world image recognition tasks. A series of extensions on CLIP through frozen models \cite{MaskCLIP,depthclip}, model fine-tuning \cite{denseclip}, visual prompts \cite{zegclip}, distillations \cite{distillingclip,3D_Distillation_clip}, and so on \cite{mask-adapted-clip} have been proposed to transfer the generalization ability of CLIP from classification to dense prediction tasks, including open-vocabulary segmentation \cite{mask-adapted-clip} and zero-shot depth estimation \cite{depthclip}. However, the feasibility of CLIP for robust restoration in low-level tasks remains unexplored. We therefore ask, is CLIP robust to image noise and can we transfer it for generalizable image denoising?    

In this paper, we find that the dense feature maps from the frozen ResNet image encoder of CLIP within specific scales exhibit remarkable resilience to noise, a property that is not easy to obtain via supervised learning. These features of clean images and their noisy counterparts show significant similarities in terms of cosine and CKA \cite{CKA} similarity measures. In addition, these features maintain a clear distinction for images with different contents and semantics. Such distortion-invariant and content-related properties are desirable for generalizable denoising as the robust and distinctive features commendably represent the latent image regardless of the corruption in the noisy measurement. 
As a result, we propose an asymmetrical encoder-decoder denoising network by integrating the frozen ResNet image encoder of CLIP and a learnable image decoder. The multi-scale features of noisy images from the frozen encoder, as well as an extra dense feature represented by the noisy image, are progressively incorporated into the decoder to recover high-quality images. Through supervised training on a single noise type and noise level, the proposed concise denoiser, termed CLIPDenoising, exhibits good generalization capacity to various OOD noises. 

By employing the frozen image encoder, the image denoising task turns into recovering clean images from fixed features. 
During training, the inherent similarity of training images \cite{chen2015external} as well as their respective dense features will inevitably affect feature diversity, leading to potential feature overfitting. Therefore, we propose progressive feature augmentation to randomly perturb these features from the frozen CLIP with increasing randomness at deeper scales. In total, our contributions are summarized as follows:

\begin{itemize}
	\item We identify that dense features from the frozen ResNet encoder of CLIP possess distortion-invariant and content-related properties. Leveraging this finding, we incorporate these features along with the noisy image into a learnable image decoder to construct a generalizable denoiser.
	\item We propose the progressive feature augmentation strategy to further improve the robustness of our approach.
	\item To the best of our knowledge, we are the first to utilize CLIP for generalizable denoising. Extensive experiments and comparisons on various OOD noises, including synthetic noise, real-world sRGB noise, and low-dose CT noise, demonstrate superior generalization of our method.
\end{itemize}

\section{Related works}

\subsection{Deep Learning-based Image Denoising}
Supervised denoising methods generally build upon powerful deep architectures (e.g., CNNs \cite{DNCNN,DRUNet}, non-local networks \cite{residualnonlocal}, and Transformer \cite{SwinIR,Restormer,StripTransformer}), large-scale paired datasets (e.g., SIDD \cite{SIDD}), and diverse optimization targets (e.g., L1/L2 losses \cite{lossl1l2}, adversarial loss \cite{gan_SR} or diffusion loss \cite{conditionaldiffusion_SR}) and have achieved state-of-the-art performance. However, the strong reliance on paired datasets and \textit{i.i.d.} assumption makes them vulnerable to unseen and OOD noise \cite{maskdenoising}. To circumvent this limitation, many unsupervised and self-supervised denoising methods \cite{Noise2void,Noise2self,Neighbor2neighbor,LIRNet,Noise2score,R2R,APBSN,CVF-SID,LGBPN} have been introduced to directly handle the target noisy images. While effective, these methods bypass tackling the generalization problem, leading to insufficient improvement in the OOD robustness of deep denoisers. Consequently, there remains a significant gap in research concerning generalizable denoising.

\subsection{OOD Generalization in Image Denoising}

Existing research on generalizable image denoising primarily consists of generalization across degradation levels and generalization across degradation types. The former addresses known noise types at unknown levels during inference, while the latter strives for general OOD robustness. Regarding the former, DnCNN \cite{DNCNN} proposed to train blind denoisers, which are capable of handling specific noise types with varying levels. Mohan et al. \cite{BFCNN} discovered that bias-free (BF) denoisers trained on limited noise ranges exhibited robustness to unseen noise levels. Consequently, the BF architecture was adopted in subsequent models like DRUNet \cite{DRUNet} and Restormer \cite{Restormer}.  Regarding the latter, GainTuning \cite{gaintuning} employed a test-time training strategy to optimize the denoiser for each noisy input. Chen et al. \cite{MeD} disentangled latent clean features from multiple corrupted versions of the same image to achieve OOD generalization.  MaskDenoising \cite{maskdenoising} revisited the commonly employed dropout operation in high-level vision tasks and integrated these units into model training.  DIL \cite{DIL} combined counterfactual distortion augmentation and meta learning-based optimization to develop a generalizable restoration network. HAT \cite{HAT} incorporated the adversarial attack and adversarial training to improve the OOD generalization of deep denoisers. Despite these advancements, there remains room for further improvement in generalizability.

\subsection{CLIP-based Generalization}
CLIP has demonstrated remarkable generalization abilities in open-world image recognition \cite{CLIP}. Subsequent works, such as MaskCLIP \cite{MaskCLIP}, DenseCLIP \cite{denseclip}, ZegCLIP \cite{zegclip}, and others \cite{mask-adapted-clip, segclip}, extended CLIP to dense prediction tasks, enabling zero-shot or open-vocabulary image segmentation. There are also studies explicitly distilling CLIP while maintaining its zero-shot performance \cite{tinyclip, distillingclip,3D_Distillation_clip}. Nevertheless, due to a distinct domain gap, there remains a lack of research on harnessing CLIP's exceptional generalization capacity for low-level vision tasks. This gap in exploration serves as a primary motivation for our work.

\subsection{Foundation Models for Image Restoration}
Utilizing foundation models to solve domain-specific tasks has become prevalent in computer vision. Similar to high-level vision tasks, many works have integrated large pre-trained models for image restoration. Diffusion model stands as a state-of-the-art generation method and many papers have leveraged the pre-learned diffusion priors to address various image restoration tasks \cite{DDPM_DDRM,score_SNIPS,Generative_Diffusion_Prior,Plug-and-Play-diffusion,cheng2023score}. Regarding Segment Anything model \cite{SAM}, some works have integrated it into image deblurring \cite{SAM_deblur}, image dehazing \cite{SAM_dehaze}, and super-resolution \cite{SAM_SR, SAM_SR2}. Yu et al. \cite{SPAE} introduced the semantic pyramid auto-encoder to enable large language models to perform image deblurring and inpainting. Additionally, Luo et al. \cite{CLIP_allinone} leveraged CLIP to predict high-quality image features and degradation features, subsequently integrating them into the image restoration model for universal image restoration. While existing works predominantly concentrate on leveraging foundation models to enhance image restoration performance, our paper aims to emphasize the enhancement of OOD generalization ability.
\section{Method}

In this section, we first check whether CLIP that has been trained on hyper-scale image-text datasets enjoys some good properties for generalizable denoising. Based on the analysis in Section \ref{Analyzing_Features}, we propose the simple and generalizable denoiser in Section \ref{simple_baseline}, followed by the strategy of progressive feature augmentation in Section \ref{Progressive_Feature_Augmentation}.

\subsection{Analyzing Features of CLIP Image Encoder} \label{Analyzing_Features}

CLIP offers two variants of image encoders, i.e., ResNet \cite{ResNet} and ViT \cite{ViT}. The ResNet version extracts multi-scale feature maps through sequential Conv-blocks and Pooling operations, while the ViT version breaks down images into smaller $16\times 16$ patches and then employs standard Transformer operations.  
For deep learning-based image denoising, low-level image details and textures are critical for reconstructing high-quality images \cite{MPRNet}. Existing methods utilize either the \textit{single-scale} pipeline \cite{SwinIR,StripTransformer} or the \textit{encoder-decoder} architecture with skip connections \cite{DRUNet,Restormer} to preserve spatial details. As the ViT architecture directly processes overly downsampled image features, it abandons spatial image details and hence is not suitable for image denoising. Consequently, we focus on analyzing and utilizing the ResNet for our further analysis and method.

\begin{figure}
	\centering
     \begin{subfigure}[b]{0.45\textwidth}
         \centering
         \includegraphics[width=\textwidth]{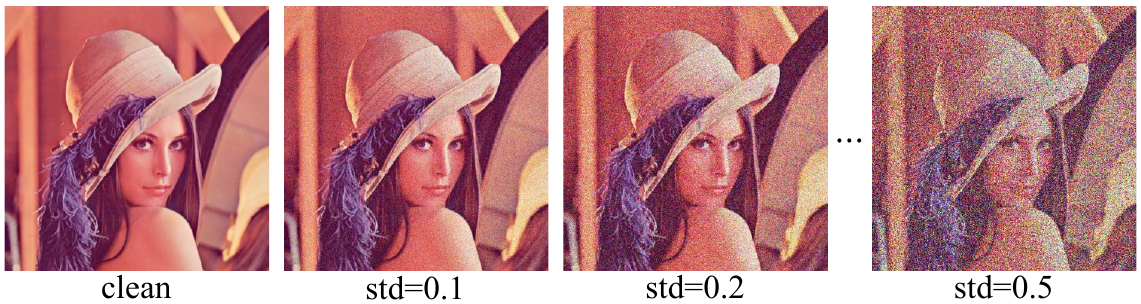}
         \caption{\textit{Lena} corrupted by different levels of $i.i.d.$ Gaussian noise}
    \end{subfigure}
	\begin{subfigure}[b]{0.47\textwidth}
         \centering
         \includegraphics[width=\textwidth]{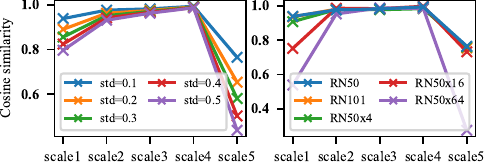}
         \caption{Left plot uses RN50 model with various noise levels while the right uses ResNets with different model sizes under std=0.1}
    \end{subfigure}
	\caption{Feature similarity analysis of the CLIP ResNet image encoder for image \textit{Lena}. Cosine similarity between $\mathbf{F}^{i}_{c}$ and $\mathbf{F}^{i}_{n}$ with regard to different noise levels and model sizes is displayed}
	\label{fig:lena_noisyfeature}
\end{figure}

\noindent \textbf{Distortion-invariant property}.  We examine the dense feature maps before each (average or attention)-pooling operation in the CLIP ResNet image encoder. (Refer Alg. \ref{alg:code} in the Supplementary Material for details). This yields a total of five multi-scale features, denoted as $\mathbf{F}^{1}\in \mathbb{R}^{ \frac{H}{2} \times \frac{W}{2} \times C}$ and $ \mathbf{F}^{i}\in \mathbb{R}^{ \frac{H}{2^i} \times \frac{W}{2^i} \times 2^{i}C}, i\in \{2, \cdots 5\}$, where $H \times W$ are the spatial dimensions of the input image, and $C$ is the base channel number. To assess the robustness of these features, we start with a clean image $\mathcal{I}_{c}$ and introduce diverse \textit{i.i.d.} Gaussian noises to create corresponding noisy images $\mathcal{I}_{n}$. Note that the image intensity range used in this section is $[0,1]$. The clean image features $\mathbf{F}^{i}_{c}$ and the noisy image features $\mathbf{F}^{i}_{n}$ are obtained by passing both the high-quality and degraded images through the frozen ResNet. Subsequently, the cosine similarity between $\mathbf{F}^{i}_{c}$ and $\mathbf{F}^{i}_{n}$ is computed at each scale $i$. We show the result in Fig. \ref{fig:lena_noisyfeature}, where five distinct noise levels and five pre-trained ResNets with increasing sizes (i.e., more residual blocks within each scale of ResNet) provided by CLIP are considered. 

\begin{figure}
    \centering
    \includegraphics{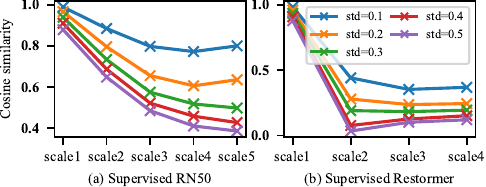}
    \caption{Feature similarity analysis of ResNet50 (supervised training for image classification, not from CLIP) and Restormer (supervised training for blind Gaussian noise removal)
    } 
    \label{fig:features_RN50_restormer}
\vspace{-3mm}
\end{figure}

From Fig. \ref{fig:lena_noisyfeature}, we observe that the initial \textit{four} features $\mathbf{F}^{i}_{n}$ from the RN50, under various corruption levels, show significant similarities to their corresponding $\mathbf{F}^{i}_{c}$, with higher similarity at deeper scales $i$. On the other hand, as the model size increases, the resemblance between $\mathbf{F}^{1}_{c}$ and $\mathbf{F}^{1}_{n}$ dramatically decreases even under smaller std=0.1. Additional results using \textit{CKA similarity} metric \cite{CKA}, \textit{Poisson degradation}, and \textit{other image} are given in Figs. \ref{fig:lena_noisyfeature_CKA}, \ref{fig:lena_noisyfeature_poisson}, \ref{fig:flowers_noisyfeature} in the Supplementary Material, which all suggest similar observations. From these findings, we conclude that features $\mathbf{F}^{i}_{n}, i\in \{1,\cdots 4\}$ from CLIP frozen RN50 are robust and distortion-invariant, which is essential for building generalizable denoisers. Additionally, we contrast these findings with feature analyses of ResNet50 trained on ImageNet for supervised image classification and Restormer trained on $i.i.d.$ Gaussian noise with $\sigma \in [0, 0.2]$ for blind denoising, and report the results in Fig. \ref{fig:features_RN50_restormer}. The distinction between Fig. \ref{fig:lena_noisyfeature} and Fig. \ref{fig:features_RN50_restormer} underlines that such distortion-invariant property is not universal and originates from CLIP. We present a brief discussion in Section \ref{discussion} about why the RN50 image encoder of CLIP holds this appealing property.

\begin{figure}
    \centering
    \begin{subfigure}[b]{0.47\textwidth}
         \centering
         \includegraphics[width=\textwidth]{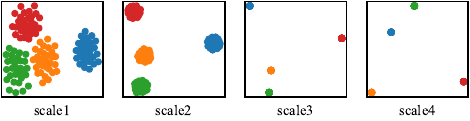}
         \caption{$\sigma=0.1$}
    \end{subfigure}
    \begin{subfigure}[b]{0.47\textwidth}
         \centering
         \includegraphics[width=\textwidth]{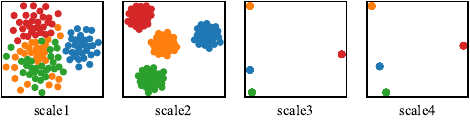}
         \caption{$\sigma=0.2$}
    \end{subfigure}
    \caption{The t-SNE plots of $\mathbf{F}^i, i \in \{1,\cdots 4\}$ from four corrupted images with diverse contents, i.e., \textit{Lena}, \textit{Baboon}, \textit{F16} and \textit{Peppers} from set9 \cite{BM3D-Set9} under $i.i.d.$ Gaussian noise with two noise levels. Different colors denote features of different images}
    \label{fig:RN50_baboon_lena_TSNE}
\end{figure}

\noindent \textbf{Conetent-related property}. 
We then check whether the above features from CLIP frozen RN50 are content-related, that is, if features of two noisy images with distinctive contents are different. Given $M$ distictive clean images $\mathcal{I}_{c}^m, m \in \{1,\cdots M\}$, we generate multiple noisy images $\mathcal{I}_{n}^m$ from $\mathcal{N}(\mathcal{I}_{c}^m, \sigma^2I)$ and obtain the corresponding multi-scale features $\mathbf{F}^{i}_{n}, i\in \{1,\cdots 4\}$ from the frozen RN50. We then compute the two-dimensional embeddings of these features using the t-SNE method \cite{t-SNE} and present the result in Fig. \ref{fig:RN50_baboon_lena_TSNE}. We find that the low-dimensional representation of $\mathbf{F}^{i}_{n}$ from different noisy image $\mathcal{I}_{n}^m$ showcases clear separation under different scale $i$ and noise level $\sigma$, indicating a strong correlation between the image content and their multi-scale features from CLIP RN50. 

\subsection{Building a Generalizable Denoiser} \label{simple_baseline}

\begin{figure}
    \centering
    \includegraphics[width=0.48\textwidth]{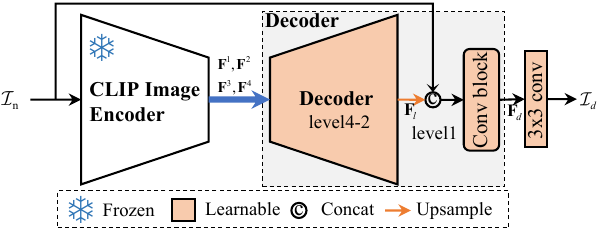}
    \caption{The CLIPDenoising for generalizable image denoising, which comprises the frozen RN50 encoder from CLIP, a learnable image decoder, and $3\times 3$ convolution}
    \label{fig:diagram}
\vspace{-2mm}
\end{figure}

Leveraging the favorable attributes of the frozen RN50 encoder from CLIP, we established a simple, effective, and generalizable denoiser, of which the architecture is depicted in Fig. \ref{fig:diagram}. Our model mainly consists of the frozen RN50 image encoder and a 4-level learnable image decoder. Given the noisy input $\mathcal{I}_n \in \mathbb{R}^{H\times W \times 3}$, the multi-scale features $\mathbf{F}^{i}_{n}, i\in \{1,\cdots 4\}$ from the frozen RN50 are first extracted. The decoder takes $\mathbf{F}^{4}_{n}$ as input and progressively recovers the high-resolution features. During the upsampling, $\mathbf{F}^{i}_{n}, i\in \{1,2,3\}$ are concatenated with the decoder features to incorporate the multi-scale feature information into the restoration (level-4 to -2). Subsequently, the decoder at level-2 outputs the feature $\mathbf{F}_l \in \mathbb{R}^{\frac{H}{2}\times \frac{W}{2} \times C}$. At level-1, $\mathbf{F}_l$ is then upsampled and concatenated with the noisy input $\mathcal{I}_n$, producing $\mathbf{F}_d \in \mathbb{R}^{H \times W \times C}$ after a final Conv-block. Ultimately, the denoised image $\mathcal{I}_d$ is obtained through a $3\times 3$ convolution operation applied on $\mathbf{F}_d$. The learnable decoder, devised to align with the ResNet encoder, is convolution-based and comprises multiple Conv-blocks, each composed of Conv-ReLU-Conv-ReLU sequences. More details about the decoder are given in the Supplementary Material. It's worth noting that the model incorporates the noisy image $\mathcal{I}_n$ as an additional image feature into the decoder, which is important and reasonable as the noisy image $\mathcal{I}_n$ itself contains rich image details and can be considered as one distinctive dense feature. This operation will be analyzed further in Section \ref{Ablations}.

During the training, we synthesize $\mathcal{I}_n$ from $\mathcal{I}_c$ based on a fixed noise type and level, and optimize the  loss function,
\begin{equation} \label{main_loss}
    \mathcal{L} = \text{E}_{p(\mathcal{I}_c)} \lVert \mathcal{I}_d - \mathcal{I}_c \rVert_1
\end{equation}

Note that, we do not employ global residual learning as we aim to restore the high-quality image from its robust features rather than restore the residual noise. In inference, we directly evaluate our model on OOD noise. Ablations in Section \ref{Ablations} indicate that our simple baseline, i.e., CLIPDenoising, has already achieved good generalizability. 

\subsection{Progressive Feature Augmentation} \label{Progressive_Feature_Augmentation}
By utilizing the frozen CLIP RN50 as the image encoder, the image denoising task turns into mapping invariant features to high-quality images. However, as images in the training dataset naturally exhibit certain degrees of similarities (e.g., similar textures in different images), the corresponding multi-scale features from CLIP RN50 tend to follow these similarities. This will reduce the feature diversity and can potentially lead to feature overfitting problems. To avoid this issue and enhance the decoder's robustness, we introduce the strategy of progressive feature augmentation, inspired by \cite{FeatAug}. During the training phase, we apply random perturbations to the multi-scale features $\mathbf{F}^{i}$ as follows
\vspace{-2mm}
\begin{equation} \label{PFA}
    \hat{\mathbf{F}}^{i} = \boldsymbol{\alpha}_i \odot \mathbf{F}^{i}, \boldsymbol{\alpha}_i \sim \mathcal{N}(\boldsymbol{1}, (\gamma \times i)^2 \boldsymbol{I}), i\in \{1,\cdots 4\}
\end{equation}
where $\odot$ denotes the element-wise multiplication and $\boldsymbol{\alpha}_i$ has the same size as $\mathbf{F}^{i}$.

In Eq. \eqref{PFA}, $\boldsymbol{\alpha}_i$ is sampled from the $i.i.d.$ Gaussian distribution with mean one and std $\gamma \times i$. For larger $i$, we inject more randomness to $\mathbf{F}^{i}$ as the deeper features tend to capture more semantic information and should also be more robust; Regarding smaller $i$, we inject less noise to preserve texture- and detail-related information contained in the shallow features. We note that such progressive feature augmentation is simple but effective.
\begin{table*}[h!]
	\centering
	\footnotesize
	\caption{Quantitative comparison (PSNR/SSIM) of different methods on CBSD68, McMaster, Kodak24 and Urban100 datasets with regard to diverse synthetic OOD noises. The best results are highlighted in \textbf{bold} and the second best is \underline{underlined}. Note that \textit{multiple noise levels} are required by HAT and DIL during the training to achieve generalization, while our method only needs \textit{one noise level} for training}
	\begin{tabular}{cccccccc}
		\hline
		Noise Types & Datasets & DnCNN \cite{DNCNN} & Restormer \cite{Restormer} & MaskDenoising \cite{maskdenoising}  &  HAT \cite{HAT} &  DIL \cite{DIL}  & Ours\\
		\hline
		\multirow{4}{*}{\shortstack{Gauss \\ $\sigma=50$}}  
		& CBSD68    & 19.84/0.363 & 19.92/0.365 & 20.68/0.432 & 20.95/0.441 & \underline{26.43/0.717} & \textbf{26.69/0.731} \\
		& McMaster       & 20.18/0.312 & 20.47/0.312 & 20.63/0.379 & 20.79/0.364 & \underline{26.61/0.669} & \textbf{27.43/0.727} \\
		& Kodak24   & 19.78/0.301 & 20.12/0.321 & 20.72/0.368 & 21.04/0.390 & \textbf{27.46/0.736} & \underline{27.39/0.723} \\
		& Urban100  & 19.62/0.420 & 19.36/0.437 & 20.51/0.485 & 20.80/0.492 & \underline{25.89/0.768} & \textbf{26.27/0.769}\\
		\hline
		\multirow{4}{*}{\shortstack{Spatial Gauss \\ $\sigma=55$}}  
		& CBSD68 & 25.91/0.699 & 23.51/0.595 & \underline{26.72/0.762} & 26.39/0.713 & 24.61/0.630 & \textbf{27.60/0.797} \\
		& McMaster & 26.18/0.649 & 24.01/0.539 & \underline{26.89/0.709} & 26.62/0.665 & 24.82/0.574 & \textbf{28.31/0.775} \\
		& Kodak24 & 25.98/0.653 & 22.99/0.533 & \underline{27.28/0.745} & 26.40/0.671 & 24.56/0.572 & \textbf{28.29/0.786} \\
		& Urban100   & 25.55/0.727  & 24.13/0.660 & 26.10/\underline{0.788} & \underline{26.48}/0.742 & 24.80/0.673 & \textbf{27.68/0.822}\\
		\hline
		\multirow{4}{*}{\shortstack{Poisson \\ $\alpha=3.5$}}  
		& CBSD68 & 24.37/0.627 & 22.20/0.559 & 24.24/0.638 & 26.61/0.733 & \underline{27.64}/\textbf{0.819} & \textbf{27.67}/\underline{0.818} \\
		& McMaster & 25.50/0.651 & 21.93/0.579 & 25.17/0.590 & 27.54/0.723 & \textbf{28.91/0.825} & \underline{28.81/0.820} \\
		& Kodak24 & 24.49/0.560 & 22.55/0.517 & 24.30/0.572 & 27.10/0.695 & \underline{28.60}/\textbf{0.821} & \textbf{28.66}/\underline{0.813} \\
		& Urban100   & 23.57/0.649 & 21.08/0.584 & 23.90/0.669 & 25.95/0.746 & \underline{27.12}/\textbf{0.854} & \textbf{27.15}/\underline{0.838}\\
		\hline
		\multirow{4}{*}{\shortstack{Salt\&Pepper \\ $d=0.02$}}  
		& CBSD68 & 26.53/0.746 & 23.59/0.679 & \underline{29.74/0.843} & 27.55/0.782 & 29.45/0.822 & \textbf{29.81/0.844} \\
		& McMaster & 25.72/0.691 & 23.05/0.640 & \underline{29.28/0.773} & 26.62/0.727 & \underline{29.28/0.773} & \textbf{29.79/0.807} \\
		& Kodak24 & 27.10/0.723 & 23.81/0.639 & \underline{30.56}/\textbf{0.842} & 28.19/0.766 & 29.99/0.810 & \textbf{30.61}/\underline{0.837} \\
		& Urban100   & 25.61/0.777 & 23.51/0.734 & 28.43/\underline{0.861} & 26.88/0.792 &\underline{29.21}/0.841 & \textbf{29.40/0.869}\\
		\hline
	\end{tabular}
	\label{tab:synthetic_result1}
\end{table*}

\section{Experiments} \label{experiments}
In this section, we first introduce experimental settings of denoising diverse OOD noises. Quantitative and qualitative results of our method and comparisons with other methods are then presented. The ablation is conducted in the last.

\subsection{Experimental Settings}

\noindent \textbf{Synthetic noise}. 
We choose $i.i.d.$ Gaussian noise with $\sigma=15$ as the in-distribution noise and consider 5 kinds of synthetic OOD noise: (1) $i.i.d$ Gaussian noise with $\sigma \in \{25, 50\}$, (2) spatial Gaussian noise with $\sigma \in \{45, 50, 55\}$, (3) Poisson noise with levels $\alpha \in\{ 2.5, 3, 3.5\}$, (4) Speckle noise with levels $\sigma^2 \in \{0.02, 0.03, 0.04\}$ and (5) Salt\&Pepper noise with levels  $d \in \{0.012, 0.016, 0.02\}$. We follow MaskDenoising \cite{maskdenoising} to generate these OOD noises and adopt Kodak24 \cite{Kodak24}, McMaster \cite{McM}, CBSD68 \cite{CBSD500}, and Urban100 \cite{Urban100} as test sets. Note that Gaussian and spatial Gaussian noises are generated in the intensity range of $[0, 255]$ while the rest uses the intensity range of $[0, 1]$, so as to be consistent with MaskDenoising.

Regarding the implementation details of our method, we build on CBSD432 dataset \cite{CBSD500} and synthesize noisy images using $i.i.d.$ Gaussian noise with $\sigma=15$ in the online fashion. Supervised training based on Eq. \eqref{main_loss} and Fig. \ref{fig:diagram} is conducted. In the training phase, we utilize the AdamW \cite{AdamM} optimizer combined with the cosine-annealing learning rate. We conduct 300k training iterations with a batch size of 16, and the learning rate decreases from an initial value of $3e^{-4}$ to a final value of $1e^{-6}$. The training patch size is $128\times 128$ and random geometric augmentations are applied to training patches. We set $\gamma=0.025$ to augment the dense features from frozen RN50 of CLIP. We perform all experiments using PyTorch \cite{paszke2019pytorch} and an Nvidia 2080ti GPU. Peak signal-to-noise ratio (PSNR) and structural similarity (SSIM) metrics are used to evaluate the denoising quality.

\noindent \textbf{Real-world sRGB noise}. For real-world sRGB noise, we consider SIDD validation dataset \cite{SIDD}, PolyU \cite{polyu} and CC \cite{CC} as test sets. These datasets comprise natural noisy sRGB images from smartphones and commercial cameras. During training, we simulate sRGB noise based on DIV2K dataset \cite{DIV2K} and the image signal processing pipeline, following CBDNet \cite{CBDNet}. In particular, we use a fixed level of Poisson-Gaussian noise, i.e., $\sigma_s=0.04, \sigma_c=0.03$ in the raw domain to generate noisy images in order to accentuate the distribution gap of data between training and testing. The optimization here is identical to that of synthetic noise.

\noindent \textbf{Low-dose CT image noise}. We utilize CLIPDenoising trained on $i.i.d$ Gaussian noise to remove real-world low-dose (LD) CT image noise, which is known to be complex and hard to model \cite{LIRNet}. We use AAPM-Mayo Clinic Low Dose CT Grand Challenge dataset \cite{AAPM_challenge}, which provides 1mm thickness abdomen slices with quarter-dose images (noisy) and corresponding normal-dose (ND) images (GT). We select 5410  NDCT images from nine patients combined with $i.i.d$ Gaussian noise with $\sigma=5$ for training and use 526 LDCT images from patient L506 for test. To accommodate the one-channel CT image, we insert a learnable $1\times 1$ convolution ahead of the CLIP RN50 encoder, which converts the single-channel image into a three-channel image. The optimization details remain the same as the above except that we set the total training iterations to 40k in this case.

\subsection{Synthetic Noise Removal}
\noindent \textbf{Compared methods}. We compare our method against three representative works on generalizable denoising, i.e., MaskDenoising \cite{maskdenoising}, DIL \cite{DIL}, and HAT \cite{HAT}. For MaskDenoising, we use the officially trained model. Regarding DIL and HAT, we follow their source codes and experimental settings to train denoisers based on $i.i.d.$ Gaussian noise with $\sigma \in \{5, 10, 15, 20\}$ and $\sigma\in[0, 25]$, respectively. Note that \textit{multiple noise levels} are required by these two methods during the training phase to achieve generalization.  We additionally evaluate DnCNN \cite{DNCNN} and Restormer \cite{Restormer} trained on $i.i.d.$ Gaussian noise with $\sigma=15$.

\begin{figure*}
\vspace{-2mm}
	\centering
	\includegraphics[width=0.90\textwidth]{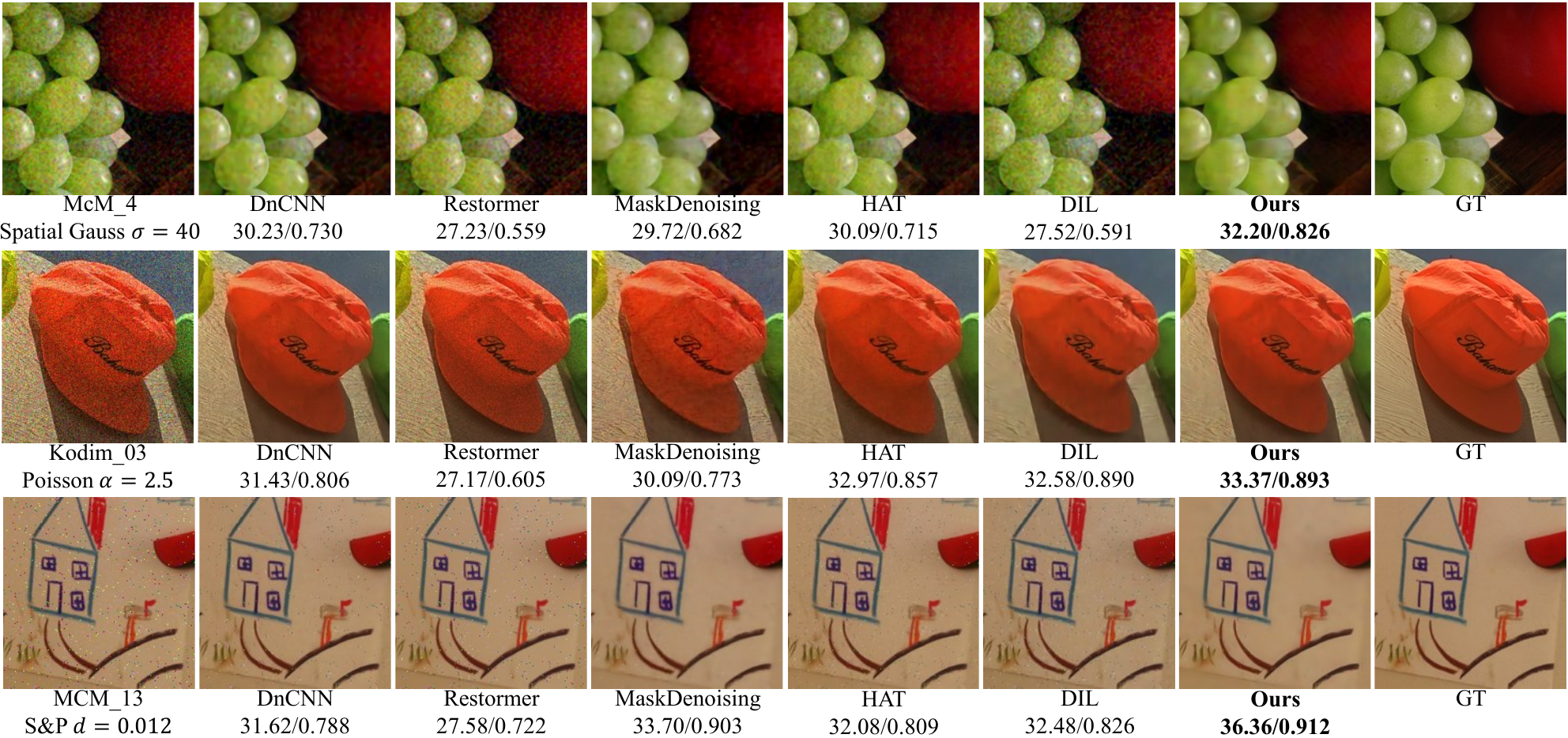}
	\caption{Qualitative denoising results on synthetic OOD noise. During the training, all the methods do not encounter the test noise types. PSNR/SSIM values are listed underneath the respective images. Zoom-in for a better comparison}
	\label{fig:synthetic_result1}
\vspace{-3mm}
\end{figure*}

\noindent \textbf{Results}. We present comprehensive quantitative comparisons of various methods across diverse noise types, levels, and datasets in Table \ref{tab:synthetic_result1} and Table \ref{tab:synthetic_result2} (in the Supplementary Material). As indicated by Tables \ref{tab:synthetic_result1} and \ref{tab:synthetic_result2}, our CLIPDenoising exhibits commendable in-distribution performance while demonstrating remarkable robustness against all considered OOD noises.  In comparison, the compared methods merely excel in some specific noise types. 

As shown in the first part of Table \ref{tab:synthetic_result2}, Restormer achieves commendable performance in in-distribution noise. However, it struggles in tackling unseen noise levels and types (see Tables \ref{tab:synthetic_result1} and \ref{tab:synthetic_result2}), indicating the overfitting to the noise in the training set. Compared with Restormer, DnCNN has a weaker modeling capacity but shows better robustness to OOD noise, which however is significantly behind the methods specialized in generalizable denoising. In comparisons among MaskDenoising, DIL, HAT, and CLIPDenoising, DIL and our method stand out as the only effective methods in eliminating Gaussian noise with higher noise level, i.e., $\sigma=50$, and our method surpasses DIL in average as indicated in Table \ref{tab:synthetic_result1}. Regarding unseen noise types, MaskDenoising outperforms HAT and DIL in spatial Gaussian noise; Conversely,  HAT and DIL exhibit notable advantages over MaskDenoising in handling Poisson and Speckle noise. In contrast, our method demonstrates consistent and competitive performance across all types of OOD noise, suggesting the great benefits from the superior distortion-invariant property of the frozen CLIP ResNet encoder. 

Note that our method, though exhibiting slightly lower performance than HAT and DIL in dealing with Speckle noise (see the middle part of Table \ref{tab:synthetic_result2}), is accomplished with just one noise level, unlike these methods which necessitate multiple noise levels during training. Fig. \ref{fig:synthetic_result1} and Figs. \ref{fig:synthetic_result2}, \ref{fig:synthetic_result3} (in the Supplementary Material) present the qualitative comparison among various methods. These visual results indicate that our method can effectively denoise OOD noise while preserving image contents and details.

\begin{figure*}
	\centering
	\includegraphics[width=0.90\textwidth]{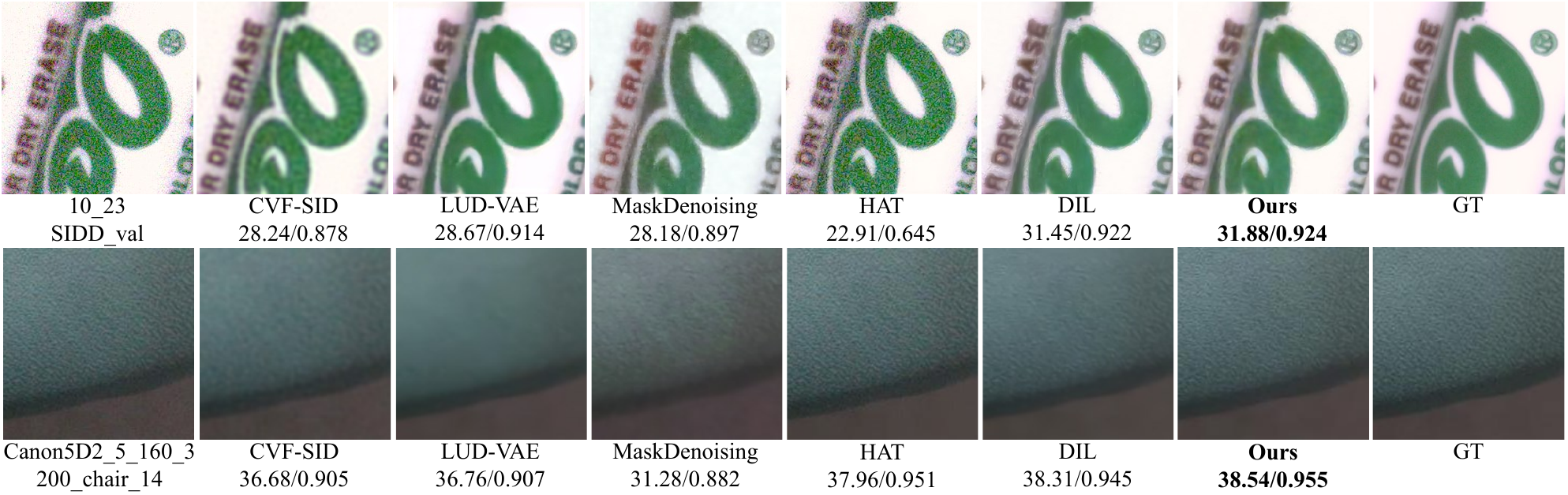}
	\caption{Qualitative results on real-world sRGB noise. PSNR/SSIM values are listed underneath the respective images.}
	\label{fig:real_result1}
 \vspace{-3mm}
\end{figure*}

\subsection{Real-world sRGB Noise Removal}
\noindent \textbf{Compared methods}. We again consider MaskDenoising, DIL, and HAT for comparison. We train MaskDenoising and DIL using their respective source codes with the same synthetic sRGB dataset used in our method. DIL mandates that the training set is categorized into four groups based on Bayer patterns. As for HAT, it's implemented based on the clean DIV2K dataset with $i.i.d.$ Gaussian noise within the range of $\sigma \in [0, 50]$ to ensure its standard functionality. In addition, we also consider two unsupervised denoising methods, i.e., CVF-SID \cite{CVF-SID} and LUD-VAE \cite{LUD_VAE}, which necessitate real-world noisy images for training.

\begin{table}
	\centering
	\footnotesize
	\caption{Quantitative comparison (PSNR/SSIM) of different methods on real-world sRGB datasets, i.e., SIDD Val, PolyU, and CC.
 }
	\begin{tabular}{cccc}
		\hline
		Methods & SIDD Val & PolyU & CC\\
		\hline
		MaskDenoising \cite{maskdenoising} & 33.14/0.913 & 24.78/0.812 & 25.63/0.881\\
		HAT \cite{HAT} & 28.58/0.570 & 37.25/0.948 & 35.27/0.901 \\
        DIL \cite{DIL} & \underline{34.76}/\underline{0.848}  & \textbf{37.65}/\underline{0.959} & \underline{36.10}/\textbf{0.948} \\
		Ours & \textbf{34.79}/\textbf{0.866} & \underline{37.54}/\textbf{0.960} & \textbf{36.30}/\underline{0.941} \\
		\hline 
		CVF-SID \cite{CVF-SID} & 34.81/0.944 & 35.86/0.937& 33.29/0.913\\
		LUD-VAE \cite{LUD_VAE} & 34.91/0.892 & 36.99/0.955 & 35.48/0.941\\
		\hline
	\end{tabular}
	\label{tab:realnoise}
 \vspace{-3mm}
\end{table}

\noindent \textbf{Results}. We present quantitative and qualitative results of different methods on real-world noise removal in Table \ref{tab:realnoise}, Fig. \ref{fig:real_result1} and Fig. \ref{fig:real_result2} (in the Supplementary Material), respectively. It is clear from Table \ref{tab:realnoise} that our method nearly achieves the best performance among the methods of generalizable denoising. Notably, DIL stands on par with our CLIPDenoising, although it requires constructing \textit{four distinct confounders}, i.e., four distortions during training, while our method does not have this requirement. In addition, compared with CVF-SID and LUD-VAE, our method shows competitive results on SIDD dataset and outperforms them on PolyU and CC datasets. As our method does not rely on real-world datasets, it emerges as a more practical and universal alternative compared to those noisy datasets-based unsupervised methods. The visual comparisons in Figs. \ref{fig:real_result1} and \ref{fig:real_result2} further underline the effectiveness of CLIPDenoising in addressing real-world sRGB noise.

\subsection{Low-dose CT Image Noise Removal}
\noindent \textbf{Compared methods}. We compare our method against the representative works in LDCT image denoising, namely Noise2Sim \cite{noise2sim} and ScoreSDE-CT \cite{score_CT}. Noise2Sim relies on the adjacent and similar LDCT images to train the denoiser, while ScoreSDE-CT learns score priors from NDCT images based on ScoreSDE and subsequently conducts posterior sampling for LDCT images. For both methods, we use their official codes and the same dataset as ours.

\begin{figure}
	\centering
	\includegraphics[width=0.45\textwidth]{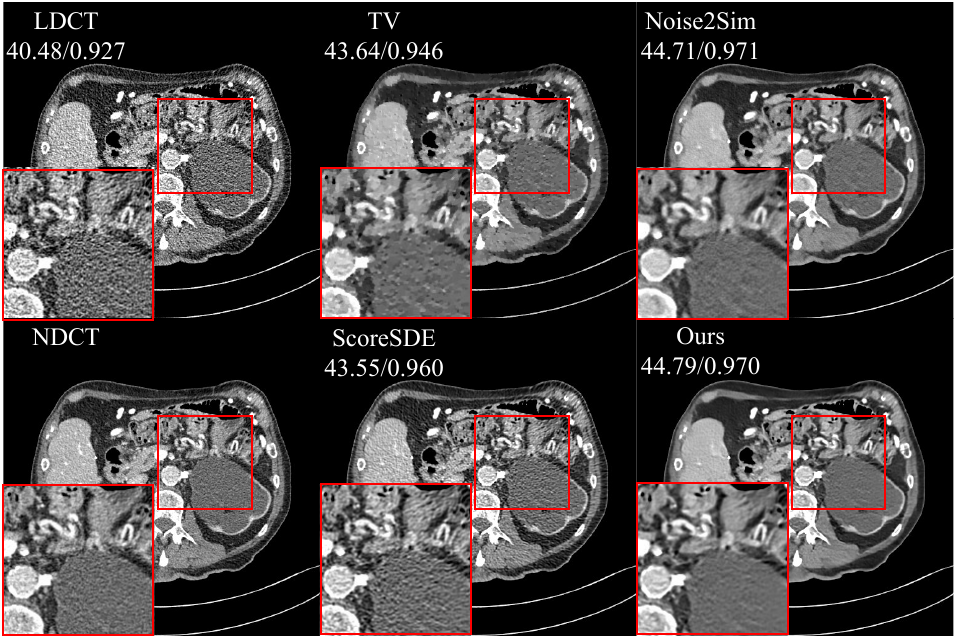}
	\caption{Denoising results on LDCT image. PSNR/SSIM are listed on the respective images. Display window: [-160, 240]HU.}
	\label{fig:CT_result}
\vspace{-2mm}
\end{figure}

\begin{table}
	\centering
	\footnotesize
	\caption{Quantitative comparison of different methods on low-dose CT abdomen slices. Note that Noise2Sim requires \textit{adjacent LDCT images} for training, while we only demand \textit{NDCT images}}
	\begin{tabular}{cccc}
		\hline
		TV & ScoreSDE \cite{score_CT} & Noise2Sim \cite{noise2sim} & Ours\\
		\hline
		44.81/0.972 & 45.36/0.972 & \textbf{45.98/0.978} & \underline{45.88/0.976}\\
		\hline
		% LDCT 42.35/0.948
	\end{tabular}
	\label{tab:ctnoise}
\vspace{-3mm}
\end{table}

\noindent \textbf{Results}. We show the quantitative outcomes in Table \ref{tab:ctnoise} and the denoised images in Fig. \ref{fig:CT_result}. Our method outperforms ScoreSDE and achieves comparable performance to Noise2Sim, indicating the effectiveness of the robust CLIP RN50 encoder in CT images. This holds substantial appeal as it enables direct knowledge transfer from models trained on natural images to medical domains, eliminating the need for, e.g., numerous adjacent low-dose CT images in Noise2Sim, especially beneficial in clinical scenarios.

\subsection{Ablations} \label{Ablations}
Here, we conduct ablation studies on synthetic noise to better evaluate the performance of our method.

\begin{table}[t] % Kodak
	\centering
	\footnotesize
	\caption{Effects of using the input image $\mathcal{I}_n$, features $\mathbf{F}^5$, different training types and progressive feature augmentation on synthetic OOD noise removal under Kodak24 dataset}
	\begin{tabular}{cccc}
		\hline
		& Gauss & Gauss & Sepckle \\ 
		&  $\sigma=15$ & $\sigma=50$ & $\sigma^2=0.04$ \\ 
		\hline
		Baseline & \underline{34.69/0.922} & \underline{26.87/0.692} & \underline{30.60/0.871} 
		\\ \hline
		w/o input $\mathcal{I}_n$ & 30.37/0.888 & 21.76/0.413 & 26.93/0.761 
		\\ 
		+ feature $\mathbf{F}^5$ & 34.03/0.916 & 25.61/0.645 & 30.39/0.865
		\\ \hline
		+ \textit{random} & 34.89/0.925 & 20.87/0.342 & 29.61/0.821 
		\\ 
		+ \textit{finetune} & 34.91/0.926 & 19.03/0.276 & 27.37/0.694 
		\\ \hline
		+ PFA & \textbf{34.69/0.922}  & \textbf{27.39/0.723} & \textbf{30.67/0.876} 
		\\ \hline
	\end{tabular}
	\label{tab:ablation1}
\vspace{-2mm}
\end{table}

\noindent \textbf{Ablation on the use of noisy image and $\mathbf{F}^5$}. Our model utilizes the noisy input as an extra dense feature to facilitate the transfer of image details and structures for restoration. Removing this operation significantly impacts the model's performance, as shown in the third row of Table \ref{tab:ablation1}. Without integrating the noisy image $\mathcal{I}_n$ into the decoder, the model's performance drastically declines, in both in-distribution and OOD cases. This highlights that although the dense features from the frozen CLIP RN50 exhibit favorable properties, they lack crucial details of the input image, rendering them ineffective for standalone restoration. In addition, utilizing feature $\mathbf{F}^5$, which does not hold distortion-invariant property as identified in Section \ref{Analyzing_Features}, also leads to a decline in both in-distribution and OOD performance.

\begin{figure}[t]
	\centering
	\includegraphics[width=0.4\textwidth]{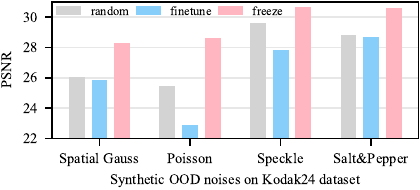}
	\caption{Robustness of \textit{random}, \textit{fintune} and \textit{freeze} (ours) to diverse OOD noises. Here, the highest OOD noise levels are used.}
	\label{fig:random_finetune_freeze}
\vspace{-5mm}
\end{figure}

\noindent \textbf{Ablation on training types}. We evaluate two optional training types to optimize our model: (1) \textit{random}, where the image encoder is randomly initialized and the entire network is trained from scratch, and (2) \textit{finetune}, which initializes the image encoder using the pre-trained RN50 from CLIP and then finetunes the complete model. As observed in Table \ref{tab:ablation1}, both training types improve the in-distribution denoising performance since the denoiser begins to overfit the training noise. Consequently, the resultant models show poor robustness to OOD noise  (also see Fig. \ref{fig:random_finetune_freeze}) as the original property of the frozen RN50 has been broken. 

\noindent \textbf{Ablation on progressive feature augmentation (PFA)}. The inclusion of progressive feature augmentation is highlighted in the last row of Table \ref{tab:ablation1}. This strategy does not affect in-distribution performance but enhances the model's generalization to OOD noise, demonstrating its efficacy.

\noindent \textbf{Ablation on ResNet versions}. In Section \ref{Analyzing_Features}, we observe that the first feature $\mathbf{F}^{1}$ from larger ResNet models of CLIP does not hold the distortion-invariant property. Here we ablate its effect on the generalization ability. We substitute the frozen RN50 model in Fig. \ref{fig:diagram} with frozen RN50x4 and RN50x16, respectively, and then train the corresponding denoisers. The outcomes, as depicted in Table \ref{tab:ablation2}, clearly indicate a substantial reduction in the generalization ability of denoisers under frozen RN50x4 and RN50x16. This underscores the significance of robust shallow features for achieving generalizable denoising.

\begin{table}
	\centering
	\footnotesize
	\caption{Effects of using different CLIP ResNet versions on synthetic OOD noise under Urban100 dataset (without PFA)}
	\begin{tabular}{cccc}
		\hline
		& Gauss & Poisson & Speckle \\
		& $\sigma=50$ & $\alpha=3.5$ & $\sigma^2=0.04$\\
		\hline
		RN50 & \textbf{25.86/0.743} & \textbf{27.11/0.837} & \textbf{28.61/0.871}\\
		RN50x4 & \underline{24.04/0.651} & \underline{26.31/0.809} & \underline{28.40/0.869} \\
		RN50x16 & 23.44/0.731& 22.26/0.762 & 18.95/0.707 \\
		\hline
	\end{tabular}
	\label{tab:ablation2}
\vspace{-3mm}
\end{table}

More experiments on the computational cost (Table \ref{tab:params_time}) and image deraining (Table \ref{tab:derain}) are provided in the supp.
\vspace{-1mm}
\section{Discussions and Limitations} \label{discussion}

\noindent \textbf{Why are dense features of CLIP RN50 encoder robust to noise?} During the training of the CLIP model, the dense features of the CLIP image decoder might overlook high-frequency information within images, including noise and fine textures. Instead, they tend to prioritize image contents to align more effectively with text semantics. This hypothesis is reasonable as the text cannot provide dense and high-frequency details for images.

\noindent \textbf{Is the robust property unique to CLIP RN50?} We hypothesize that other well-established self-supervised representation learning may induce a similar property as CLIP. To validate this, we perform feature analysis and OOD experiments using the frozen RN50 pre-trained by MoCo-v3 \cite{mocov3}, a prominent self-supervised pre-training method. The findings presented in Fig. \ref{fig:MoCo-v3-RN50} and Table \ref{tab:mocov3} (in the Supplementary Material) suggest that a denoiser based on MoCo-v3 RN50 also exhibits a certain level of OOD robustness.

\noindent \textbf{Transformer architectures for the denoiser.} As discussed in Section \ref{Analyzing_Features} and verified in Table \ref{tab:vit} of the supp, the CLIP ViT image encoder excessively downsamples image features, rendering it unsuitable for image restoration. We also attempt to implement a Transformer-based decoder by replacing the Conv-block with the Restormer-block. However, the resultant denoiser tends to overfit the training noise, leading to a decrease in its generalization. Additional strategies
are needed to mitigate this overfitting.

\vspace{-1mm}
\section{Conclusion}

This paper introduces a simple yet robust denoiser that can generalize to various OOD noises. Our method builds upon the discovery that the first four multi-scale dense features from CLIP frozen RN50 are distortion-invariant and content-related. By integrating these features and the noisy input image into the learnable image decoder, we construct a denoiser with generalization capabilities. Comprehensive experiments and comparisons on diverse OOD noises, including synthetic noise, real-world sRGB noise, and low-dose CT noise, demonstrate the superiority of our method.

\noindent \textbf{Acknowledgment}. This work was supported in part by the National Natural Science Foundation of China (NNSFC), under Grant Nos. 61672253 and 62071197.

\small
\bibliographystyle{ieeenat_fullname}
\bibliography{main}

% WARNING: do not forget to delete the supplementary pages from your submission 
\clearpage
\setcounter{page}{1}
\maketitlesupplementary

\section{More Analyses of CLIP ResNet Encoder}
\begin{algorithm}[t]
\caption{Extract dense features from CLIP ResNet.}
\label{alg:code}
\definecolor{codeblue}{rgb}{0.25,0.5,0.5}
\lstset{
  backgroundcolor=\color{white},
  basicstyle=\fontsize{7.2pt}{7.2pt}\ttfamily\selectfont,
  columns=fullflexible,
  breaklines=true,
  captionpos=b,
  commentstyle=\fontsize{7.2pt}{7.2pt}\color{codeblue},
  keywordstyle=\fontsize{7.2pt}{7.2pt},
}
\begin{lstlisting}[language=python]
# PyTorch code of CLIP ResNet image encoder
# the forward function
def forward(self, x):
    out = [] # store multi-scale dense features
    x = x.type(self.conv1.weight.dtype)
    x = self.relu1(self.bn1(self.conv1(x)))
    x = self.relu2(self.bn2(self.conv2(x)))
    x = self.relu3(self.bn3(self.conv3(x)))
    out.append(x) # scale-1 F1
    x = self.avgpool(x)
    x = self.layer1(x); out.append(x) # scale-2 F2
    x = self.layer2(x); out.append(x) # scale-3 F3
    x = self.layer3(x); out.append(x) # scale-4 F4
    x = self.layer4(x); out.append(x) # scale-5 F5
    x = self.attnpool(x)
    return out
\end{lstlisting}
\end{algorithm}

\begin{figure}[h]
    \centering
    \includegraphics{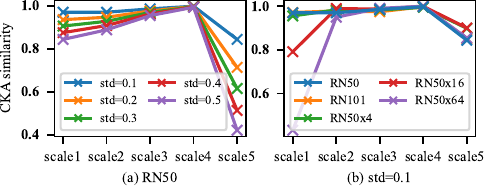}
    \caption{Feature similarity analysis of the CLIP ResNet encoder under \textit{i.i.d.} Gaussian noise with varying levels and CKA similarity measure}
    \label{fig:lena_noisyfeature_CKA}
\end{figure}

\begin{figure}[h]
    \centering
    \includegraphics{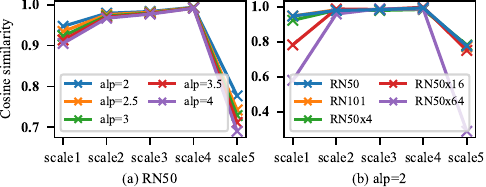}
    \caption{Feature similarity analysis of the CLIP ResNet image encoder under Poisson noise with varying levels and cosine similarity measure}
    \label{fig:lena_noisyfeature_poisson}
\end{figure}

\begin{figure}[h]
    \centering
    \includegraphics{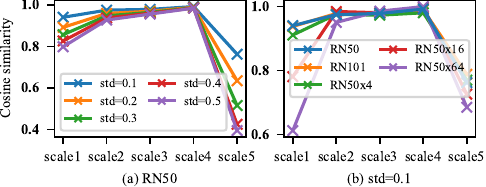}
    \caption{Feature similarity analysis for the image \textit{flowers} from set14 under Gaussian noise and cosine similarity measure}
    \label{fig:flowers_noisyfeature}
\end{figure}

We conduct more feature analysis of CLIP frozen ResNet encoder for the image \textit{Lena} using Poisson noise and CKA similarity measure, respectively, and report the results in Fig. \ref{fig:lena_noisyfeature_CKA} and \ref{fig:lena_noisyfeature_poisson}.  Besides, feature similarity analysis of CLIP ResNet encoder for the image \textit{flowers} from Set14 is also performed and shown in Fig. \ref{fig:flowers_noisyfeature}. 


\section{More Implementation Details}

\begin{figure}[h]
	\centering
	\includegraphics[width=0.35\textwidth]{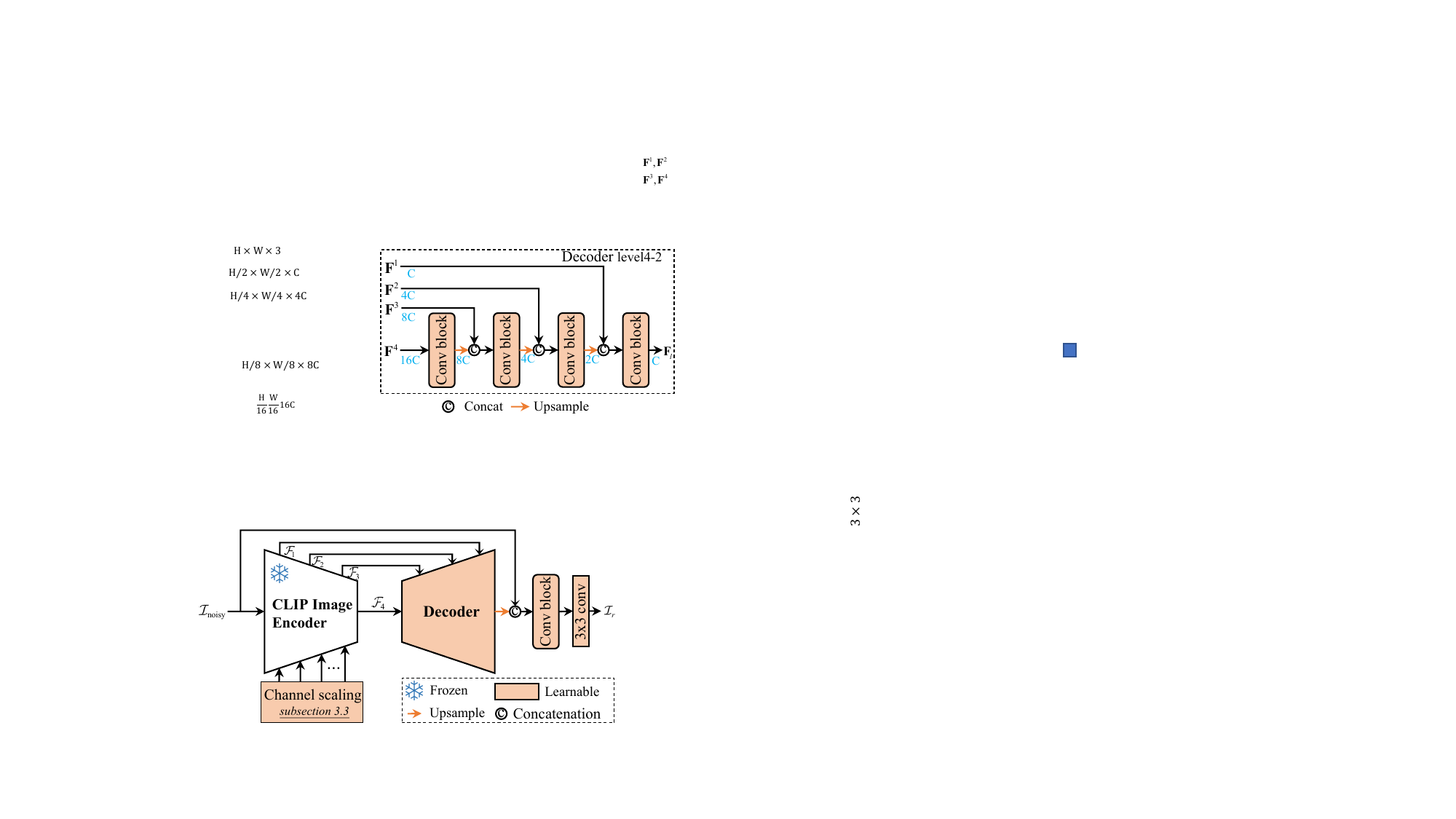}
	\caption{The detail of the learnable decoder (level-4 to -2) in our CLIPDenoising. The symbols with lightblue represent the channel number, where $C$ is the base channel number and is 64 in RN50}
	\label{fig:decoder}
\end{figure}

We present the details of our learnable decoder in Fig. \ref{fig:decoder}.

\section{More experimental results}
\subsection{Model size and inference time}
We provide the model size, inference time, and FLOPs of compared methods in Table \ref{tab:params_time}. Note that the frozen RN50 (excluding the last layer) in our model has 8.5M parameters, and the learnable decoder has 11M parameters. And HAT uses the DnCNN model.

\begin{table}[h] 
	\centering
	\footnotesize
	\caption{Efficiency comparisons (test on size $3 \times 512\times 512)$}
    \begin{tabular}{p{1.4cm}<{\centering}|p{0.6cm}<{\centering}p{0.9cm}<{\centering}p{1.2cm}<{\centering}p{0.4cm}<{\centering}p{0.4cm}<{\centering}p{0.4cm}<{\centering}}
		\hline
		 & DnCNN& Restormer & MaskDenoi. & HAT & DIL & ours \\ \hline
          Params(M) & \textbf{0.67} & 26.1 & 0.88& \textbf{0.67}& 16.6 & 19.5\\
          Infer.time(s) & 0.034 &0.415 &2.239 &0.034 & 0.867 & \textbf{0.018} \\
          FLOPs(G) & 176.1 & 564.0 & 204.9 & 176.1 & 4360 & \textbf{83.58} \\
		\hline
	\end{tabular}
	\label{tab:params_time}
\end{table}

\subsection{Results on environmental  noise}

We consider image deraining task. We use Rain100L as the training set and use \textit{Rain12}, \textit{Rain drop (S)}, and \textit{Rain and mist (S)}  as OOD test sets. Results in Table \ref{tab:derain} imply that our model exhibits better generalization capability than PromptIR (NeurIPS 2023), a recent image restoration model. 

\begin{table}[h] 
	\centering
	\footnotesize
	\caption{Results on image deraining tasks (PSNR/SSIM)}
	\begin{tabular}{c|ccc}
		\hline
         & \textit{Rain12} & \textit{Rain drop (S)} & \textit{Rain and mist (S)}\\ \hline
		PromptIR & 35.00/0.944 & 22.98/0.835 & 21.89/0.673\\ 
        ours & \textbf{35.11/0.953} & \textbf{23.66/0.838} & \textbf{28.56/0.869}
        \\
		\hline
	\end{tabular}
	\label{tab:derain}
\end{table}

\subsection{Results of using CLIP ViT image encoder}

We incorporate CLIP ViT-B/16 image encoder into our model and report the resultant results in Table \ref{tab:vit}. Specifically, the model extracts features from the middle (i.e., 7-th) layer of the frozen ViT-B/16, and then feeds it to 4 learnable ViT blocks and 4 learnable upsampling blocks that upsample the deep features back to the original image space. By comparing Table \ref{tab:vit} and Tables \ref{tab:synthetic_result1}, \ref{tab:synthetic_result2}, our model with image ViT encoder shows inferior in-distribution and OOD performance.

\begin{table}[h] % Kodak
	\centering
	\footnotesize
	\caption{Results of using CLIP ViT-B/16 image encoder}
	\begin{tabular}{c|p{1.2cm}<{\centering}p{1.2cm}<{\centering}p{1.2cm}<{\centering}p{1.2cm}<{\centering}}
		\hline
		 & Gauss(15) & Gauss(50)& Poisson(3.5) & Speckle(0.04) 
        \\
        \hline
        McM & 33.28/0.903 & 21.29/0.372 & 24.78/0.618 & 27.41/0.755 
        \\
        Kodak24 & 33.98/0.914 & 20.59/0.338 & 24.02/0.529 & 28.08/0.741 
        \\
		\hline
	\end{tabular}
	\label{tab:vit}
\end{table}

\section{Experiments on MoCo-v3 ResNet50}

\begin{figure}[h]
	\centering
	\includegraphics{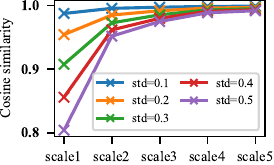}
	\caption{Feature similarity analysis of the MoCo-v3 ResNet50 under Gaussian noise. Cosine similarity is utilized here}
	\label{fig:MoCo-v3-RN50}
\end{figure}

We conduct the feature similarity analysis of frozen MoCo-v3 ResNet50 for the image \textit{Lena} using $i.i.d.$ Gaussian noise and cosine similarity measure, and report the result in Fig. \ref{fig:MoCo-v3-RN50}. Five multi-scale features show robustness to noise. Subsequently, we substitute the RN50 of CLIP with the RN50 of MoCo-v3 in our denoiser and perform the model training and OOD experiments. As observed in Table \ref{tab:mocov3}, frozen MoCo-v3 RN50-powered deep denoiser exhibits a certain level of generalization ability compared with DnCNN and our method. 

\begin{table*}[h]
	\centering
	\footnotesize
	\caption{Additional quantitative comparison of different methods on CBSD68, McMaster, Kodak24 and Urban100 datasets with regard to diverse synthetic OOD noises. The best results are highlighted in \textbf{bold} and the second best is \underline{underlined}. Note that \textit{multiple noise levels} are required by HAT and DIL during the training to achieve generalization, while our method only needs \textit{one noise level} for training}
	\begin{tabular}{cccccccc}
		\hline
		Noise Types & Datasets & DnCNN \cite{DNCNN} & Restormer \cite{Restormer} & MaskDenoising \cite{maskdenoising}  &  HAT \cite{HAT} &  DIL \cite{DIL}  & Ours\\
		\hline
		\multirow{4}{*}{\shortstack{Gauss \\ $\sigma=15$}}  
		& CBSD68 & 33.78/\underline{0.931} & \textbf{34.42/0.936} & 30.99/0.888 & 33.22/0.912 & 32.50/0.906 & \underline{33.97}/0.930 \\
		& McMaster & 34.03/\underline{0.914} & \textbf{35.61/0.935} & 30.85/0.832 & 33.04/0.883 & 32.45/0.853 & \underline{33.86}/0.910 \\
		& Kodak24 & 34.59/\underline{0.924} & \textbf{35.49/0.931} & 31.79/0.884 & 33.88/0.905 & 33.37/0.912 & \underline{34.69}/0.922 \\
		& Urban100   & 32.10/\underline{0.934} & \textbf{34.57/0.955} & 29.50/0.899 & 32.48/0.916 & 32.15/0.920 & \underline{33.15}/0.930\\		
		\hline
		\hline
		\multirow{4}{*}{\shortstack{Gauss \\ $\sigma=25$}}  
		& CBSD68 & 29.89/0.828 & 27.21/0.681 & 28.44/0.815 & \underline{30.54/0.853} & 29.98/0.843 & \textbf{31.02/0.878} \\
		& McMaster & 30.46/0.806 & 27.31/0.633 & 28.76/0.778 & \underline{30.41/0.807} & 30.17/0.790 & \textbf{31.50/0.866} \\
		& Kodak24 & 30.53/0.808 & 27.64/0.639 & 29.08/0.792 & \underline{31.37}/0.849 & 30.99/\underline{0.856} & \textbf{31.86/0.871} \\
		& Urban100   & 28.88/0.847 & 27.35/0.744 & 27.63/0.836 & \underline{29.97}/0.867 & 29.72/\underline{0.876} & \textbf{30.72/0.893}\\
		\hline
		\multirow{4}{*}{\shortstack{Spatial Gauss \\ $\sigma=45$}}  
		& CBSD68 & 28.05/0.785 & 24.51/0.668 & 28.19/\underline{0.815} & \underline{28.33}/0.791 & 26.33/0.703 & \textbf{29.43/0.849} \\
		& McMaster & 28.24/0.740 & 24.01/0.539 & \underline{28.27/0.762} & 28.44/0.742 & 26.47/0.646 & \textbf{29.82/0.825} \\
		& Kodak24 & 28.23/0.752 & 23.66/0.609 & \underline{28.85/0.805} & 28.44/0.760 & 26.31/0.652 & \textbf{30.12/0.838} \\
		& Urban100   & 27.53/0.806 & 25.80/0.722 & 27.34/\underline{0.837} & \underline{28.32}/0.809 & 26.51/0.737 & \textbf{29.42/0.868}\\
		\hline
		\multirow{4}{*}{\shortstack{Spatial Gauss \\ $\sigma=50$}}  
		& CBSD68 & 26.92/0.741 & 23.98/0.630 & \underline{27.47/0.790} & 27.32/0.752 & 25.43/0.665 & \textbf{28.51/0.825} \\
		& McMaster & 27.16/0.693 & 24.63/0.573 & \underline{27.60/0.738} & 27.49/0.703 & 25.61/0.609 & \textbf{29.10/0.803} \\
		& Kodak24 & 27.04/0.702 & 23.29/0.569 & \underline{28.09/0.778} & 27.37/0.715 & 25.39/0.611 & \textbf{29.21/0.815} \\
		& Urban100   & 26.49/0.766  & 24.93/0.690 & 26.74/\underline{0.815} & \underline{27.36}/0.776 & 25.61/0.704 & \textbf{28.55/0.847}\\
		\hline
		\multirow{4}{*}{\shortstack{Poisson \\ $\alpha=2.5$}}  
		& CBSD68 & 28.70/0.806 & 25.63/0.693 & 27.80/0.806 & \textbf{30.15}/0.858 & \underline{29.53/0.870} & \underline{29.94}/\textbf{0.874} \\
		& McMaster & 29.80/0.799 & 25.75/0.693 & 28.55/0.730 & \underline{30.87}/0.840 & 30.74/\textbf{0.871} & \textbf{30.93}/\underline{0.864} \\
		& Kodak24 & 29.23/0.776 & 26.06/0.644 & 28.42/0.781 & \underline{30.90}/0.843 & 30.42/\underline{0.868} & \textbf{30.93/0.869} \\
		& Urban100   & 27.56/0.806  & 25.14/0.719 & 26.95/0.815 & \textbf{29.52}/0.871 & 29.17/\textbf{0.893} & \underline{29.51/0.884}\\
		\hline
		\multirow{4}{*}{\shortstack{Poisson \\ $\alpha=3$}}  
		& CBSD68 & 26.37/0.712 & 23.55/0.615 & 25.87/0.718 & 28.48/0.804 & \underline{28.52/0.844} & \textbf{28.71/0.846} \\
		& McMaster & 27.49/0.720 & 23.62/0.632 & 26.13/0.667 & 29.30/0.784 & \underline{29.78}/\textbf{0.848} & \textbf{29.82}/\underline{0.843} \\
		& Kodak24 & 26.68/0.660 & 23.95/0.561 & 27.04/0.683 & 29.16/0.779 & \underline{29.45}/\textbf{0.844} & \textbf{29.71}/\underline{0.841} \\
		& Urban100   & 25.41/0.721  & 22.72/0.642 & 25.33/0.737 & 27.82/0.815 & \underline{28.09}/\textbf{0.874} & \textbf{28.27}/\underline{0.862}\\
		\hline
		\multirow{4}{*}{\shortstack{Speckle \\ $\sigma^2=0.02$}}  
		& CBSD68 & 31.79/0.898 & 29.10/0.826 & 29.91/0.875 & \textbf{32.50}/\underline{0.916} & 31.57/\textbf{0.924} & \underline{31.82}/0.904 \\
		& McMaster & 32.74/0.886 & 28.89/0.800 & 30.47/0.809 & \textbf{33.11}/\underline{0.899} & \underline{32.66}/\textbf{0.907} & 32.28/0.870 \\
		& Kodak24 & 32.82/0.895 & 29.96/0.814 & 30.80/0.874& \textbf{33.26}/\underline{0.908} & 32.35/\textbf{0.919} & \underline{32.91}/\underline{0.908} \\
		& Urban100   & 30.11/0.893  & 28.24/0.828 & 28.60/0.883 & \textbf{31.49}/\underline{0.917} & 30.90/\textbf{0.930} & \underline{30.94}/0.904\\
		\hline
		\multirow{4}{*}{\shortstack{Speckle \\ $\sigma^2=0.03$}}  
		& CBSD68 & 30.10/0.856 & 26.78/0.765 & 28.99/0.851 & \textbf{31.10}/\underline{0.893} & 30.40/\textbf{0.906} & \underline{30.48}/0.886 \\
		& McMaster & 31.21/0.846 & 26.81/0.752 & 29.70/0.778 & \textbf{31.95}/\underline{0.873} & \underline{31.70}/\textbf{0.904} & 31.31/0.858 \\
		& Kodak24 & 31.12/0.852 & 27.50/0.740 & 29.90/0.848 & \textbf{31.95}/0.884 & 31.28/\textbf{0.901} & \underline{31.64/0.891} \\
		& Urban100   & 28.37/0.841  & 25.86/0.774 & 27.65/0.847 & \textbf{30.10}/\underline{0.892} & \underline{29.72}/\textbf{0.916}  & 29.69/0.889\\
		\hline
		\multirow{4}{*}{\shortstack{Speckle \\ $\sigma^2=0.04$}}  
		& CBSD68 & 28.65/0.812 & 25.13/0.719 & 27.94/0.815 & \textbf{29.97}/0.867 & \underline{29.56}/\textbf{0.890} & 29.49/\underline{0.870} \\
		& McMaster & 29.69/0.804 & 25.30/0.717 & 28.68/0.736 & \underline{30.90/0.845} & \textbf{30.94/0.893} & 30.47/\underline{0.845} \\
		& Kodak24 & 29.53/0.801 & 25.66/0.683 & 28.76/0.804 & \textbf{30.82}/0.856 & 30.49/\textbf{0.887} & \underline{30.67/0.876} \\
		& Urban100   & 26.92/0.795  & 24.17/0732 & 26.64/0.804 & \textbf{28.86}/0.862 & \underline{28.82}/\textbf{0.903} & 28.69/\underline{0.875}\\
		\hline
		\multirow{4}{*}{\shortstack{Salt\&Pepper \\ $\alpha=0.012$}}  
		& CBSD68 & 28.56/0.814 & 25.88/0.779 & 30.49/0.863 & 29.31/0.846 & \underline{30.81/0.865} & \textbf{31.95/0.890} \\
		& McMaster & 27.76/0.773 & 25.32/0.746 & 30.11/0.798 & 28.39/0.804 & \underline{30.44/0.820} & \textbf{31.90/0.863} \\
		& Kodak24 & 29.17/0.797 & 26.17/0.751 & \underline{31.27/0.861} & 29.91/0.834 & 31.24/0.851 & \textbf{32.72/0.882} \\
		& Urban100   & 27.40/0.823  & 25.73/0.815 & 29.08/\underline{0.880} & 28.60/0.851 & \underline{30.49}/0.875 & \textbf{31.50/0.901}\\
		\hline
		\multirow{4}{*}{\shortstack{Salt\&Pepper \\ $\alpha=0.016$}}  
		& CBSD68 & 27.45/0.780 & 24.57/0.726 & \underline{30.13/0.853} & 28.32/0.813 & 30.02/0.841 & \textbf{30.85/0.857} \\
		& McMaster & 26.61/0.730 & 24.00/0.687 & 29.70/0.786 & 27.37/0.763 & \underline{29.75/0.793} & \textbf{30.85/0.838} \\
		& Kodak24 & 28.05/0.760 & 24.81/0.690 & \underline{30.94/0.853} & 28.95/0.797 & 30.52/0.827 & \textbf{31.67/0.863} \\
		& Urban100   & 26.42/0.806  & 24.45/0.771 & 28.76/\underline{0.871} & 27.63/0.820 & \underline{29.75}/0.855  & \textbf{30.30/0.889}\\
		\hline
	\end{tabular}
	\label{tab:synthetic_result2}
\end{table*}

\begin{table*}[h]
	\centering
	\footnotesize
	\caption{Quantitative comparison of DnCNN, our model with frozen CLIP RN50, and our model with frozen MoCo-v3 RN50 on McMaster and Kodak24 datasets with regard to various synthetic OOD noises. All methods are trained under $i.i.d.$ Gaussian noise with $\sigma=15$. Progressive feature augmentation is not used here.}
	\begin{tabular}{cccccc}
		\hline
		McMaster& Gauss $\sigma=50$ & Spatial Gauss $\sigma=55$ & Poisson $\alpha=3.5$ & Speckle $\sigma^2=0.04$  & S\&P $d=0.02$  \\
		\hline
		DnCNN             & 20.18/0.312 & 26.18/0.649 & 25.50/0.651 & \underline{29.69}/0.804 & 25.72/0.691 \\
		Ours+CLIP RN50    & \textbf{26.95/0.698} & \textbf{28.24/0.771} & \textbf{28.82/0.814} & \textbf{30.29/0.824} & \underline{29.62/0.795} \\
		Ours+MoCo-v3 RN50 & \underline{24.85/0.625} & \underline{27.36/0.748} & \underline{26.92/0.747} & 29.68/\underline{0.823} & \textbf{29.96/0.809}  \\
		\hline
		\hline
		Kodak24& Gauss $\sigma=50$ & Spatial Gauss $\sigma=55$ & Poisson $\alpha=3.5$ & Speckle $\sigma^2=0.04$  & S\&P $d=0.02$  \\
		\hline
		DnCNN             & 19.78/0.301 & 25.98/0.653 & 24.49/0.560 & 29.53/0.801 & 27.10/0.723\\
		Ours+CLIP RN50    & \textbf{26.87/0.692} & \textbf{28.19/0.781} & \textbf{29.74/0.840} & \textbf{30.60/0.871} & \underline{30.52/0.832}\\
		Ours+MoCo-v3 RN50 & \underline{25.59/0.630} & \underline{27.72/0.747} & \underline{26.82/0.718} & \underline{29.90/0.829} & \textbf{30.95/0.834}\\
		\hline
	\end{tabular}
	\label{tab:mocov3}
\end{table*}

\begin{figure*}
	\centering
	\includegraphics[width=0.88\textwidth]{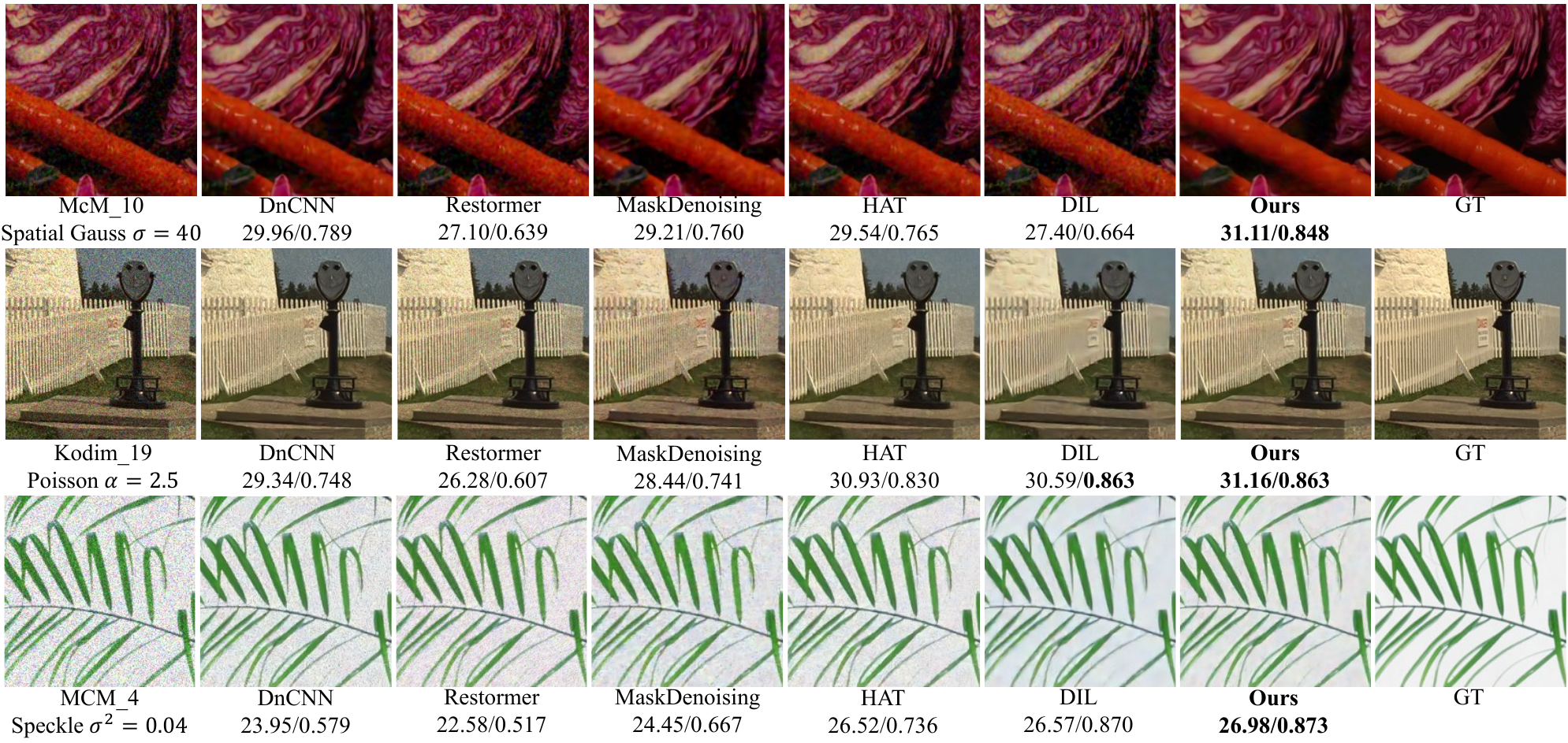}
	\caption{More qualitative denoising results on synthetic OOD noise.}
	\label{fig:synthetic_result2}
\end{figure*}

\begin{figure*}
	\centering
	\includegraphics[width=0.88\textwidth]{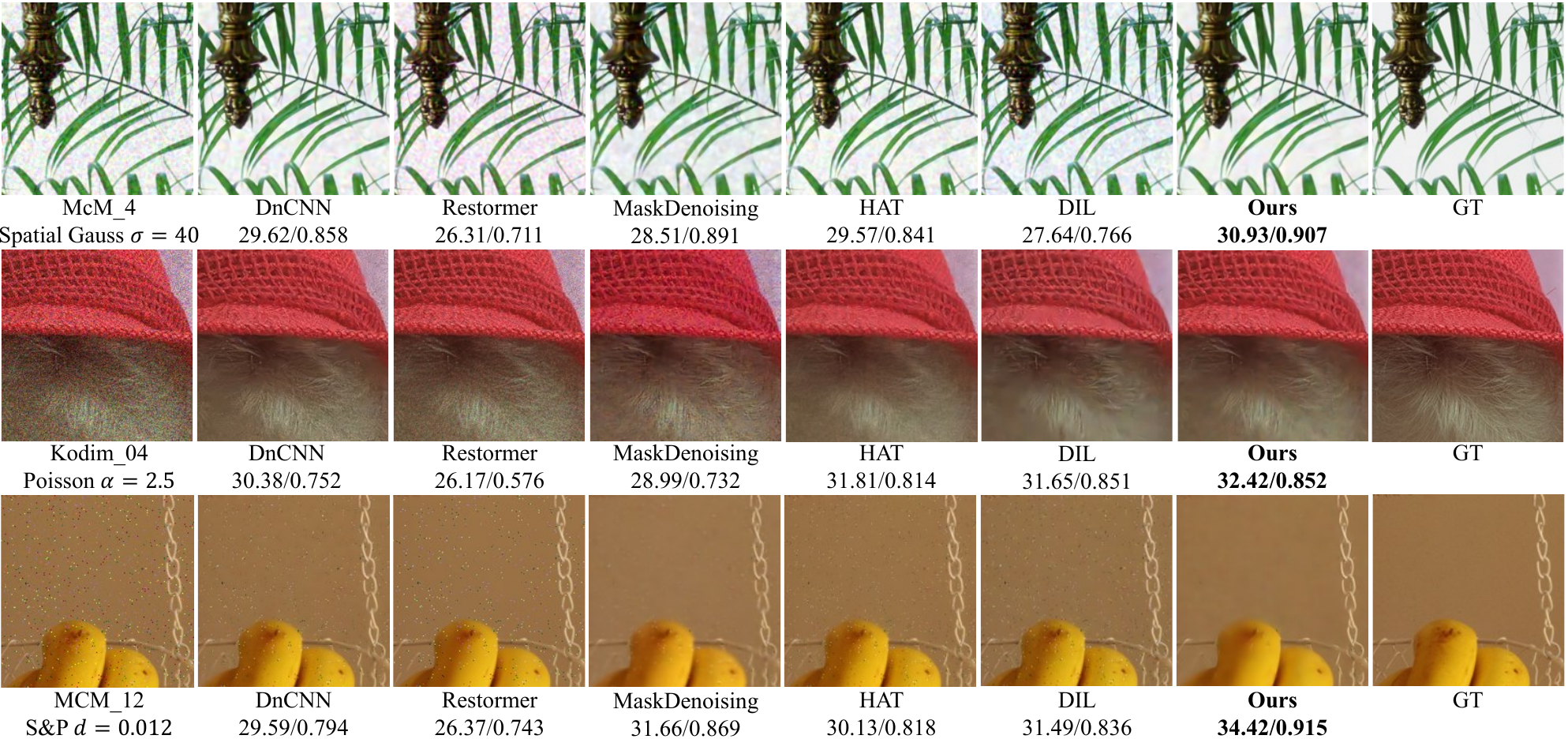}
	\caption{More qualitative denoising results on synthetic OOD noise.}
	\label{fig:synthetic_result3}
\end{figure*}

\begin{figure*}
	\centering
	\includegraphics[width=0.88\textwidth]{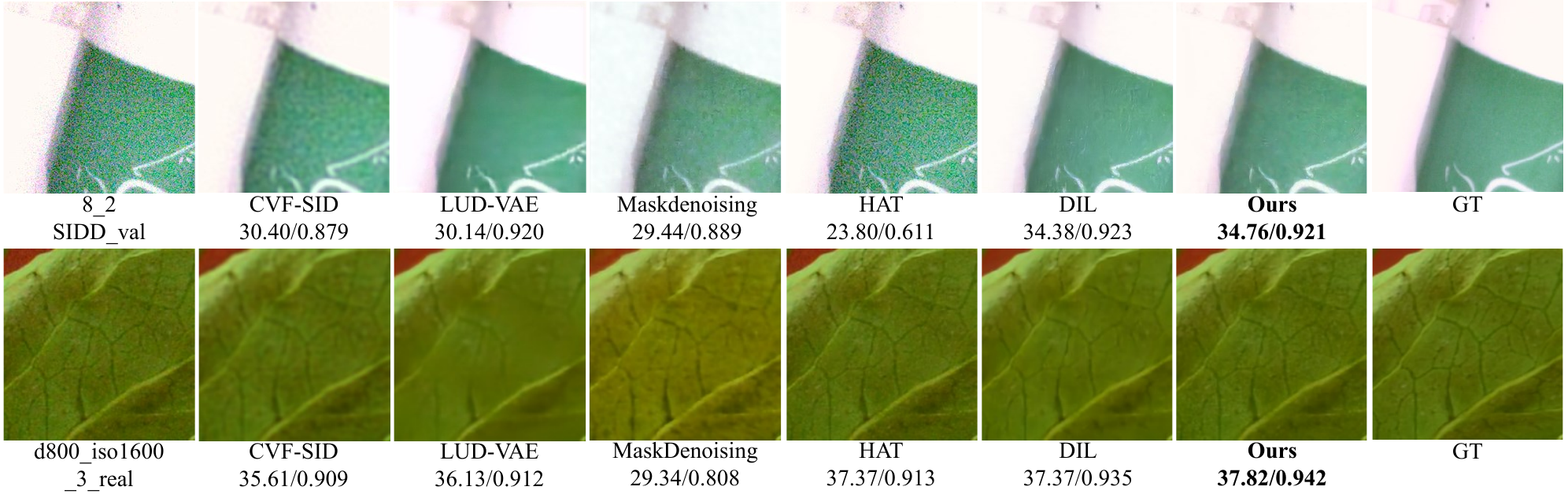}
	\caption{More qualitative denoising results on real-world sRGB noise}
	\label{fig:real_result2}
\end{figure*}

\end{document}